\documentclass[twoside]{article}

\usepackage[accepted]{arxiv}
\usepackage[numbers]{natbib}
\renewcommand{\bibname}{References}
\renewcommand{\bibsection}{\subsubsection*{\bibname}}

\usepackage{amsmath}
\usepackage{amsfonts}
\usepackage{amsthm}
\usepackage{amssymb}
\usepackage{color}
\usepackage{multirow}

\usepackage{booktabs}
\usepackage{graphicx}
\usepackage{xcolor}
\usepackage{apxproof}
\usepackage{appendix}
\usepackage[normalem]{ulem}

\newtheoremrep{thm}{Theorem}
\newtheoremrep{prop}{Proposition}
\newtheoremrep{lem}{Lemma}
\newtheoremrep{obs}{Observation}
\DeclareMathOperator*{\argmax}{arg\,max}

\renewcommand{\appendixprelim}{\clearpage 
\twocolumn[

\aistatstitle{Distilling Policy Distillation [Appendix]}

\aistatsauthor{Wojciech Marian Czarnecki \And  Razvan Pascanu \And  Simon Osindero} 
\aistatsaddress{DeepMind \And DeepMind \And DeepMind}
\aistatsauthor{Siddhant Jayakumar \And  Grzegorz M. \'{S}wirszcz \And Max Jaderberg}
\aistatsaddress{DeepMind \And DeepMind \And DeepMind}

] }

\begin{document}
\runningauthor{Wojciech Marian Czarnecki et al.}

\begin{toappendix}

\setcounter{section}{0}
\section{Experimental details}

\subsection{MDPs}

The MDPs used in this study are $W \times H$ grid worlds, meaning that the state space is $\mathcal{S} = \{ s_{i,j} \}_{i,j=1}^{W,H} \cup \{ s_{term}\}$. $s_\text{term}$ is a special state, to which an agent is moved with probability 0.01 after each action, ensuring finite length of the experiments considered. There is one initial state placed in the centre of the grid, $\mathcal{S}_1 = \{ s_{\lceil W/2 \rceil, \lceil H/2 \rceil} \}$. 
There are four possible actions $\{L, R, U, D\}$, each of them has an associated desired effect, namely $L(s_{i,j}) = s_{i-1,j}$, $R(s_{i,j}) = s_{i+1,j}$, $D(s_{i,j}) = s_{i,j-1}$, $U(s_{i,j}) = s_{i,j+1}$. Some transitions are invalid, as they would lead to leaving the state space, thus we define 
$$
z(s_{i,j}, a) = \left \{
\begin{matrix}
a(s_{i,j}) \;\;\; \text{if } a(s_{i,j}) \in \mathcal{S}\\
s_{i,j} \;\;\; \text{otherwise}
\end{matrix}
\right . 
$$
The transition dynamics are defined as:
$$
T(a, s_{i,j}) = \left \{
\begin{matrix}
z(s_{i,j}, a) \;\;\; \text{with probability } 1-\eta, \\
z(s_{i,j}, L) \;\;\; \text{with probability } \eta/4, \\
z(s_{i,j}, R) \;\;\; \text{with probability } \eta/4, \\
z(s_{i,j}, U) \;\;\; \text{with probability } \eta/4, \\
z(s_{i,j}, D) \;\;\; \text{with probability } \eta/4, \\
\end{matrix}
\right .
$$
where $\eta = 0.1$ is the transition noise.

Rewards are associated with some states, and are fully deterministic.

Some states are terminal, which cause the episode to end, and bring the agent back to the initial state.

We considered partial, and fully observable versions of these environments. In the fully observable environments, the agent is given the state index as an observation, while in the partially observable environments a concatenated sequence of $(2k+1) \times (2k+1)$ objects, namely $o_k(s_{i,j})$  is represented as
\begin{equation*}
\begin{aligned}
(&o(s_{i-k,j-k}), ..., o(s_{i-k, j}), ...,o(s_{i-k, j+k}), ..., \\
&o(s_{i-k+1, j-k}), ..., o(s_{i-k+1, j}), ...,o(s_{i-k+1, j+k}), ..., \\
&...\\
&o(s_{i+k, j-k}), ..., o(s_{i+k, j}), ...,o(s_{i+k, j+k}))
\end{aligned}
\end{equation*}
where $o(s)$ is $\emph{wall}$ if $s \notin \mathcal{S}$ and a pair $(\mathrm{reward\_value}(s), \mathrm{is\_terminating}(s))$ otherwise. For example, if the state provides reward 10 and is terminating, then it will be observed as \emph{(10, True)}.

In all partially observable experiments, we use observations which are concatenations of $9 \times 9$ squares of vision, centered in an agent position. 
We experimented with visual extents ranging from $5 \times 5$ to full observability and found that this does not effect the qualitative results of the paper, thus the choice of the particular visual extent is not crucial.

\begin{figure}
\centering
\includegraphics[width=0.5\textwidth]{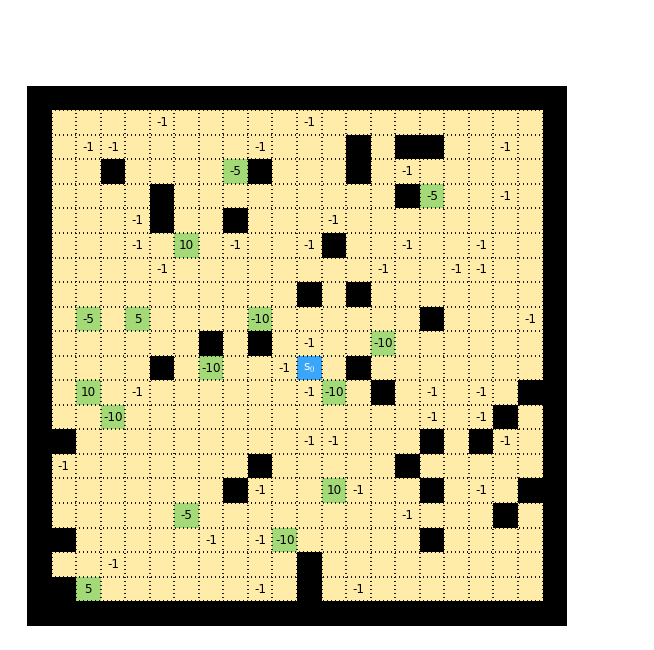}
\caption{Example 20$\times$20 grid world MDP, with initial state coloured blue, terminal states coloured green, and rewards on various states. Black squares are \emph{wall}s.}
\label{fig:grid}
\end{figure}

\subsection{Distribution over MDPs}

In all experiments where we sample multiple MDPs we use the following procedure:

\begin{enumerate}
    \item We create $\mathcal{S}$ as described in the previous section.
    \item For each $s_{ij} \in \mathcal{S}$, starting in the upper left corner and traversing first horizontally and then vertically:
    \begin{enumerate}
        \item With probability $p_{w}$ we remove $s_{ij}$ from $\mathcal{S}$, which we call putting a wall in; if we modified a state, we go back to step 2 and continue the loop.
        \item With probability $p_{+10}$ we put a reward of +10 in $s_{ij}$ and make it terminal; if we modified a state, we go back to step 2 and continue the loop.
        \item With probability $p_{+5}$ we put a reward of +5 in $s_{ij}$ and make it terminal; if we modified a state, we go back to step 2 and continue the loop.
        \item With probability $p_{-1}$ we put a reward of -1 in $s_{ij}$; if we modified a state, we go back to step 2 and continue the loop.
        \item With probability $p_{-5}$ we put a reward of -5 in $s_{ij}$ and make it terminal; if we modified a state, we go back to step 2 and continue the loop.
        \item With probability $p_{-10}$ we put a reward of -10 in $s_{ij}$ and make it terminal; if we modified a state, we go back to step 2 and continue the loop.
    \end{enumerate}
    \item We check if there exists a path between the initial state and the $+10$ state, and if this is not true, we repeat the process.

\end{enumerate}

Unless otherwise stated in the text, we use $W=H=20$, $\tfrac{1}{10} p_w = p_{+10} = \tfrac{1}{2} p_{+5} = \tfrac{1}{10} p_{-1} = p_{-5} = p_{-10} = 0.01$.

\subsection{Actor Critic}

We use a basic actor critic method, where we sample one full episode under the student policy, $\tau \sim \pi_\theta$, and then update the parameters according to either the single sample Monte Carlo estimated return:
$$
\nabla_\theta \log \pi_\theta(a_t | \tau_t) [ \sum_t \gamma^{t-1} r_t - V_\theta(s_t) ]
$$
or, in the TD(1) case, with bootstrapped estimates
$$
\nabla_\theta \log \pi_\theta(a_t | \tau_t) [ r_t + \gamma V_\theta(s_{t+1}) - V_\theta(s_t) ].
$$
In all experiments we used $\gamma=0.99$, but we obtained qualitatively similar results with other values too ($\gamma=0.95$ and $\gamma=0.999$).

After each update we use the same Monte Carlo or TD value to fit the baseline function, using the L$^2$ loss:
$$
\nabla_\theta (V_\theta(s_t) - \gamma^{t-1} r_t)^2
$$or$$
\nabla_\theta (V_\theta(s_t) - (r_t + \gamma V_\theta(s_{t+1}))^2
$$
in the case of TD learning, where $V_\theta(s_{t+1})$ is treated as a constant.
All Vs are initialised to 0s. The learning rate used is $0.1$.

\subsection{Q-Learning}

We use the standard Q-Learning update rule of
$$
Q(a_t,s_t) := (1-\lambda) Q(a_t,s_t) + \lambda ( r_t + \gamma \max_a Q(a,s_{t+1} ))
$$
applied after each visited state. All Qs are initialised to 0s. The learning rate was set to $\lambda = 0.01$. The policy was trained for 30k iterations.

When treating the Q-Learned policy as a teacher, depending on the temperature $T$ reported (by default 0) it was either a greedy policy (if the temperature is 0) 
$$
\hat \pi(a|s) = 1 \text{ iff } Q(a,s) = \max_{b \in \mathcal{A}} Q(b,s)
$$
$$
 \pi(a|s) = \tfrac{\hat \pi(a|s)}{\sum_{b\in \mathcal{A}}\hat \pi(b|s)}$$
or a Boltzman policy computed as:
$$
\pi(a|s) = \tfrac{\exp(Q(a,s)/T)}{\sum_{b \in \mathcal{A}} \exp(Q(b,s)/T)}
$$

\subsection{Policy parametrisation during distillation}

Policies are represented as logits of each action, for each unique observation. Consequently for each observation $o$, and for action space $\mathcal{A}$ the policy for Actor Critic is parameterised as $\pi_\theta(a|o) = \frac{\exp(\theta_{a,o})}{\sum_{b \in \mathcal{A}} \exp( \theta_{b,o} )}$.

Similarly, value functions are represented simply as one float per observation: $V_{\pi_\theta}(o) = \theta^V_{o}$, and Q-values $Q_{\pi_\theta}(a,o) = \theta^Q_{a,o}$.

\section{Extended figures}

We include extended versions of various figures. Fig.~\ref{fig:big_comp_ext} is an extended version of Fig.~\ref{fig:big_comp} including experiments with an A2C teacher.
\begin{figure*}[h]
    \centering
    \includegraphics[width=\textwidth]{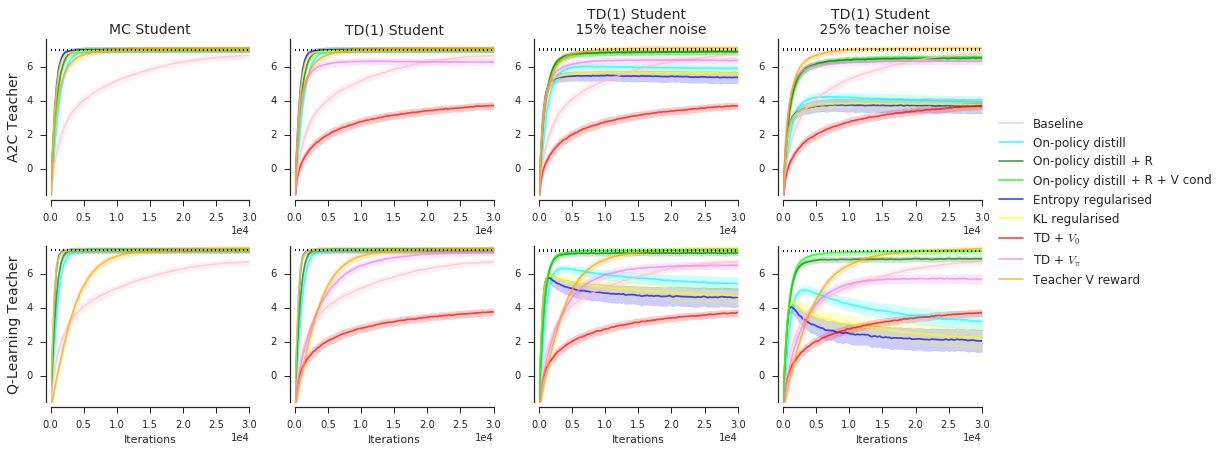}
    \caption{Learning curves obtained from averaging 1k training runs on randomly generated MDP grid worlds. The gradual decrease in reward when distilling from a sub-optimal Q-Learning teacher with 
    distillation methods that enforce full \emph{policy cloning} comes from the fact that the teacher is purely deterministic -- while being closer to it initially helps, once the student replicates all the wrong state decisions perfectly its reward start to decrease. Extended version of Fig.~\ref{fig:big_comp} including A2C teacher.}
    \label{fig:big_comp_ext}
\end{figure*}

Fig.\ref{fig:mdp_v_example_ext} is an extended version of Fig.~\ref{fig:mdp_v_example}, including more sizes of the corridor environment.
\begin{figure*}[htb]
    \centering
    \includegraphics[width=\textwidth]{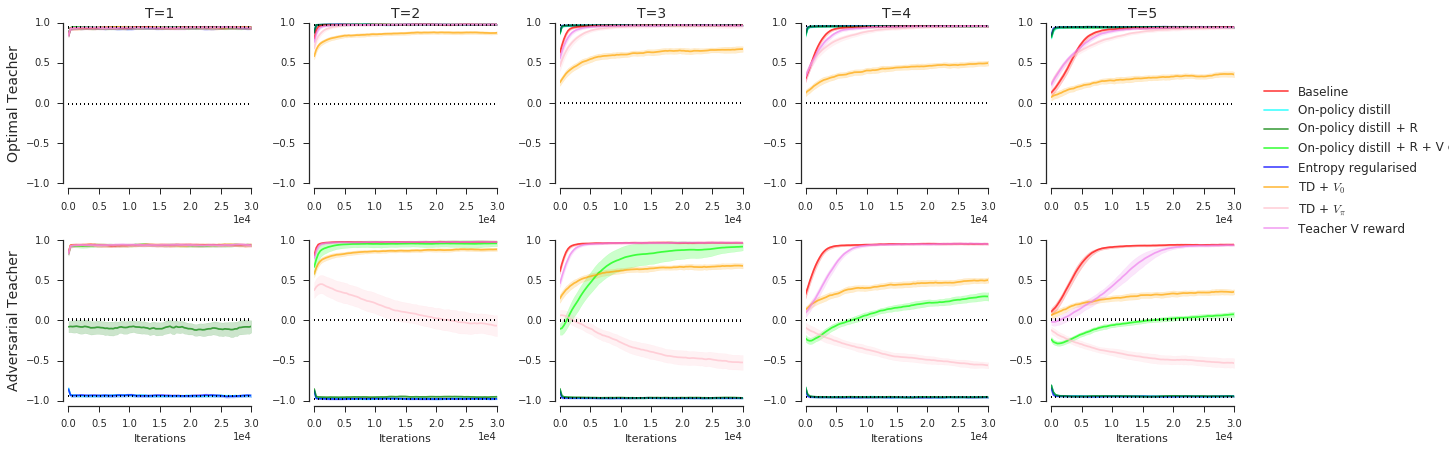}
    \includegraphics[width=\textwidth]{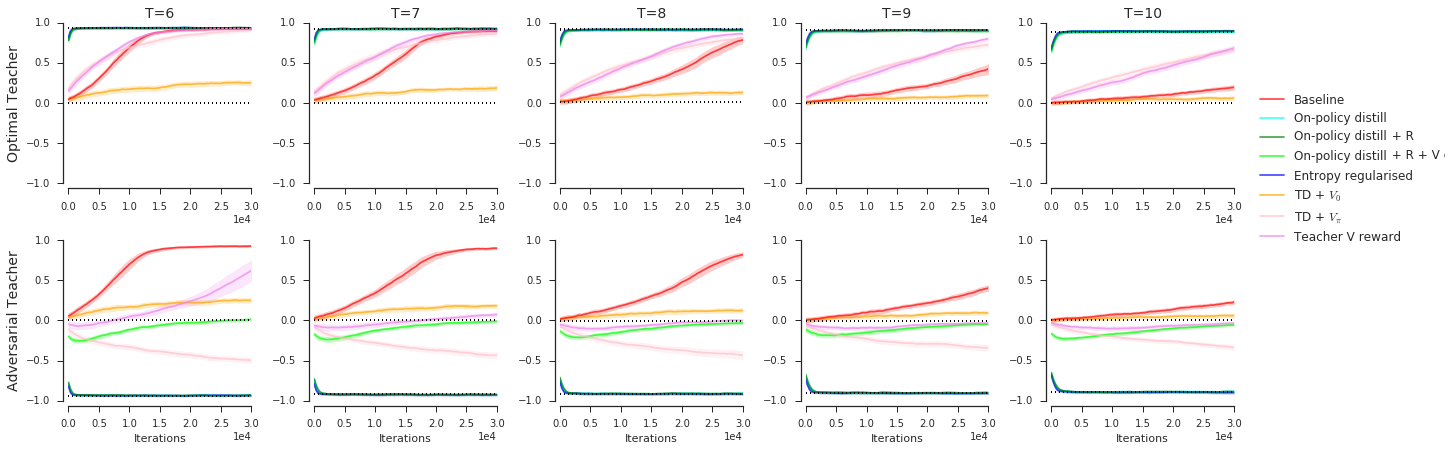}
    \caption{Results of distilling the optimal teacher and adversarial (minimising reward) teacher in the chain-structured MDP. Extended version of Fig.~\ref{fig:mdp_v_example}.}
    \label{fig:mdp_v_example_ext}
\end{figure*}

Fig.\ref{fig:actionspace_ext} is an extended version of Fig.~\ref{fig:actionspace}, including additional agents.
\begin{figure*}[htb]
\centering
\includegraphics[width=\textwidth]{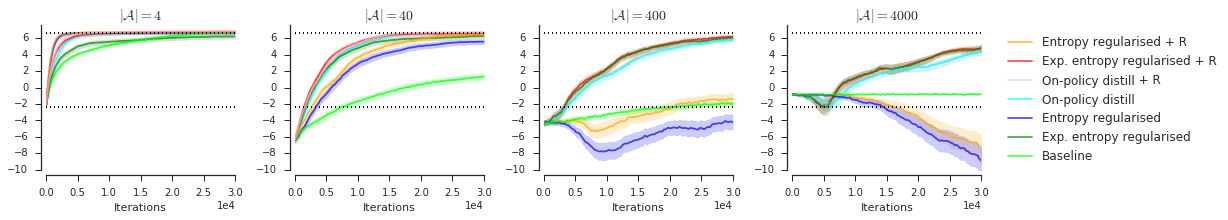}
\caption{Learning curves averaged over 1k random MDPs with $|\mathcal{A}|$ actions, out of which 4 are movement actions and the remaining ones do not affect the movement of the agent, but simply make exploration hard. Plots show the failure mode of the intrinsic reward only based distillation, and how their expected version fixes it. Extended version of Fig.~\ref{fig:actionspace}}
\label{fig:actionspace_ext}
\end{figure*}

\end{toappendix}

\twocolumn[

\aistatstitle{Distilling Policy Distillation}

\aistatsauthor{Wojciech Marian Czarnecki \And  Razvan Pascanu \And  Simon Osindero} 
\aistatsaddress{DeepMind \And DeepMind \And DeepMind}
\aistatsauthor{Siddhant Jayakumar \And  Grzegorz \'{S}wirszcz \And Max Jaderberg}
\aistatsaddress{DeepMind \And DeepMind \And DeepMind}

]

\begin{abstract}

The transfer of knowledge from one policy to another is an important tool in Deep Reinforcement Learning. This process, referred to as {\it distillation}, has been used to great success, for example, by enhancing the optimisation of agents, leading to stronger performance faster, on harder domains~\cite{schmitt2018kickstarting,teh2017distral,czarnecki2018mix,gangwani2017genetic}.
Despite the widespread use and conceptual simplicity of distillation, many different formulations are used in practice, and the subtle variations between them can often drastically change the performance and the resulting objective that is being optimised.
In this work, we rigorously explore the entire landscape of policy distillation, comparing the motivations and strengths of each variant through theoretical and empirical analysis.
Our results point to three distillation techniques, that are preferred depending on specifics of the task. 
Specifically a newly proposed \emph{expected entropy regularised distillation} allows for quicker learning in a wide range of situations, while still guaranteeing convergence.
\end{abstract}

\begin{figure*}[htb]
\centering
\includegraphics[width=0.9\textwidth]{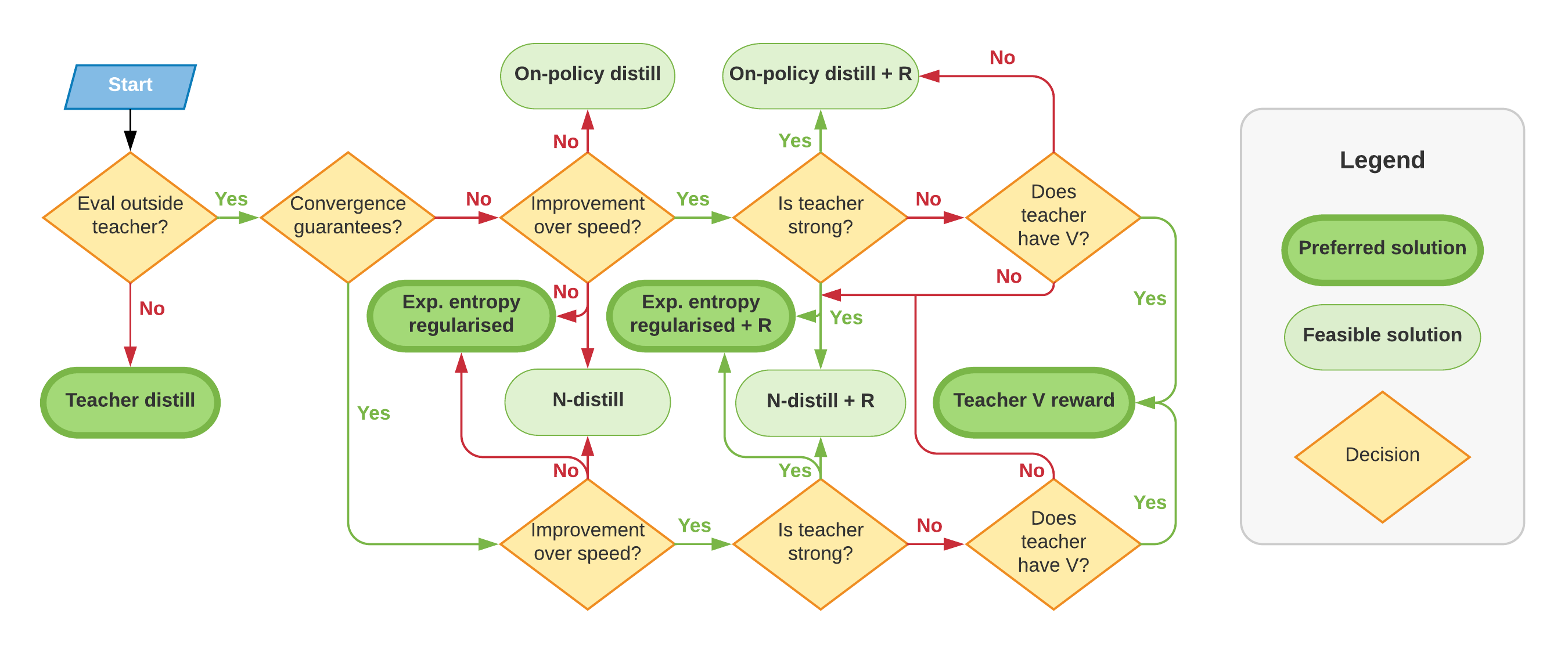}
\caption{A distillation method selection diagram, based on the results from this paper. Note, that in practise there might be many other factors affecting such decisions, but this serves as a good initial rule of thumb.
Method references:
\textbf{Teacher distill}:~\cite{rusu2015policy,ross2011reduction,ghosh2017divide};
\textbf{On-policy distill}:~\cite{ross2011reduction,parisotto2015actor,lin2017collaborative};
\textbf{On-policy distill+R}:~\cite{schmitt2018kickstarting,czarnecki2018mix}; 
\textbf{Entropy regularised}:~\cite{teh2017distral,schulman2017equivalence}; 
\textbf{N-distill}: Section~\ref{sec:control_policy_math};
\textbf{Exp. entropy regularised}: Section~\ref{sec:control_loss_emp}; 
\textbf{Teacher V reward}: Section~\ref{sec:int_reward}.
See Table~\ref{tab:my_label} for methods equations and properties.
}
\label{fig:main}
\end{figure*}

\section{Introduction}

Reinforcement Learning (RL) and in particular Deep Reinforcement Learning (DRL) has shown great success in recent years, allowing us to train agents capable of playing Atari games from raw-pixel inputs~\citep{mnih2015human, espeholt2018impala}, beating professional players in the game of GO~\citep{silver2016mastering} or learning dexterous robotic manipulation~\citep{openaidexterous}.
However, obtaining these levels of performance can often require an almost prohibitively large amount of experience to be acquired by the agent in order to learn~\citep{salimans2017evolution}.
Consequently, there is great interest in techniques that can allow for knowledge transfer; that is enabling the training of agents based on already trained policies~\citep{rusu2015policy} 
or human examples~\citep{abbeel2004apprenticeship}. 
One of the most successful techniques for knowledge transfer is that of \emph{distillation} \cite{hinton2015distilling,rusu2015policy,ross2011reduction}, where an agent is trained to match the state-dependent probability distribution over actions provided by a teacher.
Some examples include allowing us to: train otherwise untrainable agent  architectures~\citep{czarnecki2018mix}; speed up learning~\citep{schmitt2018kickstarting}; build stronger policies~\citep{ross2011reduction}; drive multi-task learning~\citep{arora2018multi,rusu2015policy}.
Analogous techniques have been widely used in supervised learning problems to achieve model compression~\citep{hinton2015distilling,howard2017mobilenets, polino2018model}, reparametrisation for inference speedup~\citep{oord2017parallel}, and joint co-training of multiple networks~\citep{zhang2017deep}.

Although the high-level formulation of distillation in RL is simple, one can find dozens of different mathematical formulations used in practice. For example: sometimes trajectories are sampled from a teacher~\citep{rusu2015policy}, sometimes from the student~\citep{lin2017collaborative} or a mixture~\citep{ross2011reduction}; some authors use a KL divergence between teacher and student distribution~\citep{ghosh2017divide} while others look at KL between entire trajectory probabilities~\citep{teh2017distral}. 
A primary goal of this paper is to provide a roadmap  of these different ideas and approaches, and then to perform a step-by-step comparison of them, both mathematically and empirically. 
This allows us to construct a set of useful guidelines to follow when trying to decide which specific distillation approach might best fit a particular problem.

The main contributions of this paper can be summarised as follows: 
In Section~\ref{sec:control_policy_math} we provide a proof that 
commonly used distillation with trajectories sampled from the student policy does not form a gradient vector field, and while it has convergence guarantees in simple tabular cases, it can oscillate as soon as one introduces rewards to the system. We show simple methods of recovering the gradient vector field property.
In Section~\ref{sec:control_policy_emp} we perform empirical evaluation of different control policies, showing when and why it is beneficial to use student-driven distillation. 
In Section~\ref{sec:actor_critic} we analyse the actor-critic setup, in which one has access to a teacher's value function, $V$, in addition to its policy, and discuss how $V$ may also be used for distillation.
We empirically evaluate all the above techniques in thousands of random MDPs.
Finally, based on all the results combined from our mathematical analyses and empirical evaluation, we propose effective new distillation variants and provide a rule of thumb decision tree, Fig.~\ref{fig:main}.
\section{Preliminaries}
Throughout the paper we assume that we are working with Markov Decision Processes, and will now outline our notation. 
We have a finite set of states, $\mathcal{S}$, a finite set of possible actions, $\mathcal{A}$, an agent policy $\pi : \mathcal{S} \rightarrow \Delta^{|\mathcal{A}|}$ which outputs distribution over actions in each state. 
Agents interact with an environment at time $t$ by sampling actions $a_t \sim \pi(\tau_t)$, and the environment transitions to a new state according to unknown transition dynamics $\tau_{t+1} \sim T(\tau_t, a_t)$ and produces rewards $r_t \sim r(\tau_t, a_t)$. Each $\tau_t \in \mathcal{S}$ is the state encountered at time $t$ in the trajectory $\tau$.
We use $\mathbb{E}_\pi$ to denote an expectation over the distribution of trajectories, $\tau$, generated by agent when interacting with the environment using policy $\pi$. 
Under this notation the typical goal of reinforcement learning is to find $\pi^* = \argmax_{\pi}{\left[\mathbb{E}_\pi[ \sum_{t=1}^{|\tau|} \gamma^{t-1}r_t]\right]}$, where $\gamma \in [0,1]$ is discount factor. 
For simplicity we use $\gamma=1$ in most of our theoretical results, though the proofs can trivially be extended to arbitrary $\gamma \leq 1$. For the empirical results we generally use $\gamma = .99$, see the Appendix A.
We consider the general problem of extracting knowledge from a \emph{teacher policy}, $\pi$, and transferring it to a different \emph{student policy}, $\pi_\theta$, using trajectories, {$\tau = \{\tau_1, a_1, r_1, \ldots, \tau_{|\tau|}, a_{|\tau|}, r_{|\tau|}\}$}, sampled from interactions between a control policy, $q_\theta$, and the unknown environment.

All proofs are provided in the Appendices C and D.

\begin{table*}[htb]
    \centering
   \caption{A comparison of various ways of defining distillation between the teacher ($\pi$) and student ($\pi_\theta$) policies. $q_\theta$ is the sampling (control) policy. $\nabla$ column denotes whether the update rule prescribed is a valid gradient vector field. 
    $\text{H}^\times(p(s)\|q(s))$ denotes Shannon's cross entropy between two distributions over actions $-\mathbb{E}_{a\sim p(s)} \left [ \log q(a|s) \right ].$
    * For proof see Theorem 1. ** For proof see Theorem 2.
    Methods below the mid line are introduced in this paper -- usually as modifications of known techniques to address specific issues identified. Each of the techniques but \emph{Teacher distill} and \emph{Teacher V reward} has a corresponding \emph{+R} version where $\widehat r_i$ is replaced with $\widehat r_i + r_i$. }
    \begin{tabular}{lllllll}
    \toprule
 name & $q_\theta$ &  $\ell(\pi_\theta, V_{\pi_\theta}, \tau_{t})$ &  $  \widehat r_i$
 & is  $\nabla$? & Loss\\
 \midrule
Teacher distill & $\pi$ & $\text{H}^\times(\pi(\tau_{t}) \| \pi_\theta(\tau_{t}))$ & 0 & yes~[1] & \scriptsize $\mathbb{E}_\pi [ \sum_t \text{H}^\times(\pi(\tau_t) \| \pi_\theta(\tau_t)) ]$ \\
On-policy distill & $\pi_\theta$& $\text{H}^\times(\pi(\tau_{t}) \| \pi_\theta(\tau_{t}))$&0 & no$^{*}$ & does not exist$^{*}$ \\
Entropy regularised& $\pi_\theta$ & 0 & $\log \pi(a_i|\tau_i)$ & yes~[4] &\scriptsize $\mathbb{E}_{\pi_\theta} [ \sum_t -\log \pi(a_t|\tau_t) ]$\\
\midrule
N-distill & $\pi_\theta$& $\text{H}^\times(\pi(\tau_t) \| \pi_\theta(\tau_t))$& -$\text{H}^\times(\pi(\tau_{i+1}) \| \pi_\theta(\tau_{i+1}))$ & yes$^{**}$ &\scriptsize $\mathbb{E}_{\pi_\theta} [ \sum_t \text{H}^\times(\pi(\tau_t) \| \pi_\theta(\tau_t)) ]$\\
Exp. entropy regularised& $\pi_\theta$ &  $\text{H}^\times(\pi_\theta(\tau_{t}) \| \pi(\tau_{t}))$ & $\log \pi(a_{i+1}|\tau_{i+1})$ & yes$^{**}$ & \scriptsize $\mathbb{E}_{\pi_\theta} [ \sum_t -\log \pi(a_t|\tau_t) ]$\\
Teacher V reward & $\pi_\theta$ & 0 & $r_i +V_\pi(\tau_{i+1}) - V_{\pi_\theta}(\tau_i)$ & yes$^{**}$ &\scriptsize $\mathbb{E}_{\pi_\theta} [ \sum_t r_t ]$\\
         \bottomrule
    \end{tabular}
     
    \label{tab:my_label}
\end{table*}
\section{Policy distillations}

Through the rest of this paper we consider update rules for $\theta$ (parameters of the student policy $\pi_\theta$) which are proportional to:
\begin{equation}
\begin{aligned}
\mathbb{E}_{q_\theta}  \left [ \sum_{t=1}^{|\tau|}
 - \nabla_\theta \log \pi_\theta(a_t | \tau_t) 
\widehat {R_t}
+
\nabla_\theta \ell(\pi_\theta, V_{\pi_\theta}, \tau_t) \right ],
\end{aligned}
\label{eqn:update_rule}
\end{equation}
for
$\widehat {R_t} = \sum_{i=t}^{|\tau|} \widehat r_i = \sum_{i=t}^{|\tau|} \widehat r(\pi_\theta, V_{\pi_\theta}, \tau_{i}, a_{i}, \tau_{i+1}, a_{i+1}, r_{i})$
and a choice of $q, \ell$ and $\hat r$ that define a specific instance of a distillation technique (see Table~\ref{tab:my_label} for a list of examples). 
In this equation, $\ell$ can be seen as a form of auxiliary loss~\citep{jaderberg2016reinforcement} responsible for policy alignment at the current step, while $\hat r$ can be viewed as a reward term that combines extrinsic and intrinsic components~\citep{pathak2017curiosity} and thus is responsible for long-term alignment.
Note, we assume undiscounted objectives and episodic RL, but analogous analysis can be performed for the discounted case.

Focusing on update rules rather than simply losses may seem to add unnecessary complexity, however one of the crucial outcomes of our work is to show that the update rules involved in certain distillation methods do not have corresponding loss functions. Consequently we must make an explicit distinction between 
\emph{update rules}, and \emph{losses which may be used to derive update rules}.

\subsection{Control policy}
\label{sec:control_policy_math}
Many RL distillation frameworks set up knowledge transfer as a supervised learning problem~\citep{ross2011reduction,rusu2015policy,ghosh2017divide,yin2017knowledge}, by following updates in the direction of:
$
\mathbb{E}_\pi  \left [ \sum_{t=1}^{|\tau|} \nabla_\theta \text{H}^\times(\pi(\tau_t)\|\pi_\theta(\tau_t)) \right],
$
with Monte Carlo estimates for the expectation based on trajectories derived from the teacher policy, $\pi$. However, since then, several publications~\citep{parisotto2015actor,czarnecki2018mix,schmitt2018kickstarting} have reported better empirical results when trajectories are sampled from the student instead, i.e. by following updates in the direction of:
$
\mathbb{E}_{\pi_\theta} \left [ \sum_{t=1}^{|\tau|} \nabla_\theta \text{H}^\times(\pi(\tau_t)\|\pi_\theta(\tau_t)) \right].
$
Note that in this form of update the gradient operator is under an expectation wrt. the same set of variables that it operates upon. Consequently it is not clear if this process will even converge, and thus the benefits of using such updates are also unclear.

In this section we analyse and prove the following properties:
    (i) For tabular policies, provided $q_\theta$ guarantees a non-zero probability of sampling each state visited by the teacher, the dynamics will converge. In particular $q_\theta = \pi_\theta$ with a softmax policy satisfies this property;
    (ii) In general, updates like this do not form gradient vector fields;
    (iii) If one adds reward optimisation to the system, the dynamics can cycle and never converge;
    (iv) A reward-based correction term can be added to ensure convergence (and with such a correction, the updates do correspond to proper gradient vector field);
    (v) There is a trade-off between the speed of convergence and the fidelity of the behaviour replication, which can be controlled by $q_\theta$.

We begin by proving a general theorem about on-policy non-gradients. In principle, it is very similar to the notion of \emph{compatibility} of a value function and the policy~\cite{sutton2000policy}, or can be seen as a generalisation of \emph{incompatibility} towards 
other possible trajectory level losses $\ell(\tau|\theta)$ (e.g. $\ell(\tau|\theta)=\sum_{t=1}^{|\tau|} \ell(\pi(\tau_t)\| \pi_\theta(\tau_t))$).

\begin{thmrep}
Let us assume that $g(\theta) = \mathbb{E}_{\pi_\theta} [\nabla_\theta 
\ell(\tau|\theta)]$
is differentiable
and there does not exist $\alpha_\tau \in \mathbb{R}$ such that 
$\nabla_\theta \ell(\tau | \theta) =  \alpha_\tau \nabla \pi_\theta(\tau) $ almost everywhere.
Then $g(\theta)$ is not a gradient vector field of any function.
\label{theorem:not_gvf}
\end{thmrep}
\begin{appendixproof}
If gradient of some $f$ is differentiable then $f$'s Hessian exists and is a symmetric matrix:
$$
\frac{\partial}{\partial x} \left (\frac{\partial}{\partial y} f(x, y) \right )
=
\frac{\partial}{\partial y} \left (\frac{\partial}{\partial x} f(x, y) \right ).
$$
Consequently, if some function is a gradient vector field, then its Jacobian has to be symmetric.
We will show that for $g$ this is not true in general, by focusing on two arbitrary indices $ij$ and $ji$. We use notation $f[i]$ to denote the $i$th output of the multivariate function $f$. Using the log derivative trick we obtain that $\tfrac{\partial}{\partial \theta_j} g(\theta)[i]$ equals
\begin{equation*}
\begin{aligned}
 &
&&\tfrac{\partial}{\partial \theta_j}\mathbb{E}_{\pi_\theta} \left [ \tfrac{\partial}{\partial \theta_i} \log \pi_\theta(\tau) \ell(\tau, \theta) \right ] \\
&= 
&&\mathbb{E}_{\pi_\theta} \left [ \tfrac{\partial}{\partial \theta_j} \log \pi_\theta(\tau) \tfrac{\partial}{\partial \theta_i} \log \pi_\theta(\tau) \ell(\tau, \theta) \right ]\\
&&&+
\mathbb{E}_{\pi_\theta} \left [ \tfrac{\partial}{\partial \theta_j} [ \tfrac{\partial}{\partial \theta_i} \log \pi_\theta(\tau) \ell(\tau, \theta) ] \right ]\\
&= 
&&\mathbb{E}_{\pi_\theta} \left [ \tfrac{\partial}{\partial \theta_j} \log \pi_\theta(\tau) \tfrac{\partial}{\partial \theta_i} \log \pi_\theta(\tau) \ell(\tau, \theta) \right ]\\
&&&+
\mathbb{E}_{\pi_\theta} \left [  \tfrac{\partial}{\partial \theta_i \theta_j} \log \pi_\theta(\tau) \ell(\tau, \theta) \right ]\\
&&&+
\mathbb{E}_{\pi_\theta} \left [  \tfrac{\partial}{\partial \theta_i} \log \pi_\theta(\tau) \tfrac{\partial}{\partial \theta_j} \ell(\tau, \theta)  \right ]
\end{aligned}    
\end{equation*}
thus $
\tfrac{\partial}{\partial \theta_i} g(\theta)[j] - 
\tfrac{\partial}{\partial \theta_j} g(\theta)[i]$ equals
\begin{equation*}
\begin{aligned}
 \mathbb{E}_{\pi_\theta}&  \left [ 
\tfrac{\partial}{\partial \theta_j } \log \pi_\theta(\tau)  
\tfrac{\partial}{\partial \theta_i} \ell(\tau, \theta) -
\tfrac{\partial}{\partial \theta_i } \log \pi_\theta(\tau)  
\tfrac{\partial}{\partial \theta_j} \ell(\tau, \theta)
\right ]\\
& = \int_\tau  \left [
\tfrac{\partial}{\partial \theta_j } \pi_\theta(\tau)  
\tfrac{\partial}{\partial \theta_i} \ell(\tau, \theta) -
\tfrac{\partial}{\partial \theta_i } \pi_\theta(\tau)  
\tfrac{\partial}{\partial \theta_j} \ell(\tau, \theta)
\right ] d\tau
\end{aligned}
\end{equation*}
In general this term is zero iff  $\nabla \ell(\tau, \theta) = 
\alpha_\tau \nabla \pi_\theta(\tau)$ almost everywhere,
which can not be true due to assumptions. Consequently, $g(\theta)$ is not a gradient vector field of any function. 
\end{appendixproof}

The assumption about the existence of $\alpha_\tau$ is equivalent to the compatibility criterion~\citep{sutton2000policy}, and thus it shows that incompatible value functions do not create valid gradient vector fields. This is complementary to the result that compatible value functions provide convergence to the optimal policy.
\begin{figure*}[htb]
    \centering
    \includegraphics[width=0.32\textwidth]{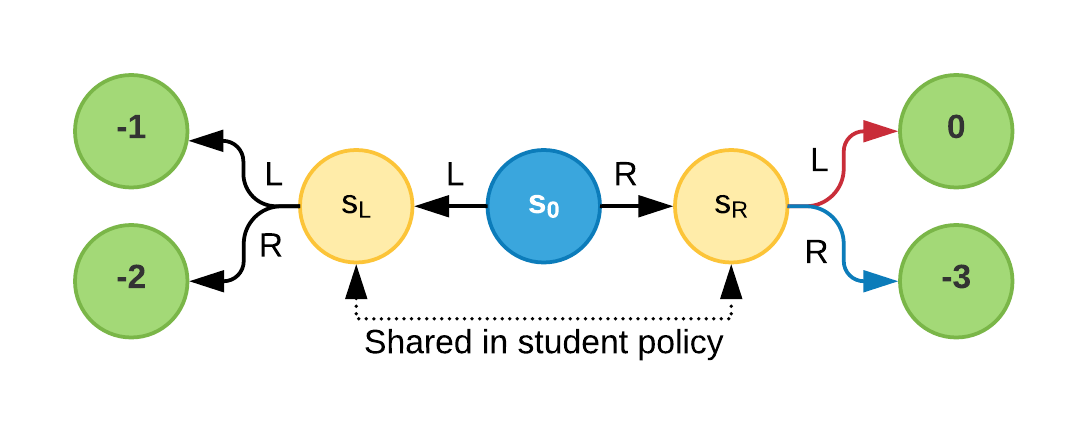}
    \includegraphics[width=0.33\textwidth]{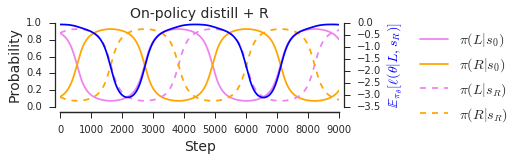}
    \includegraphics[width=0.33\textwidth]{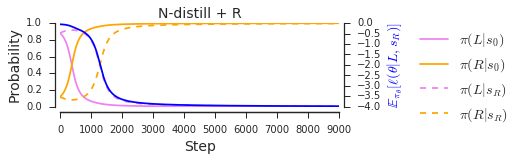}
    \caption{\textbf{Left:} An MDP providing an example of non-convergence of on-policy distillation, when distilling from a teacher which prefers to go right (blue arrow) when in $s_R$ over going left (red arrow) by putting a loss $\ell(\theta|s_R) = -4\pi_\theta(R|s_R)$. For simplicity it does not provide any learning signal in the remaining states. The initial state is coloured blue, L and R are actions, and green states are terminal states with rewards inside. \textbf{Center and Right:} Evolution of the policy is shown, with clear oscillation for on-policy distill and convergence for n-distill. The student policy is parameterised with sigmoids, shares parameter for $s_L$ and $s_R$, and trained with true expected returns.
    Detailed proof of oscillations is provided in the Appendix C. 
    }
    \label{fig:counter}
\end{figure*}
If we choose $\ell(\tau|\theta) =  \sum_{t=1}^{|\tau|}  \text{H}^\times(\pi(\tau_t)\|\pi_\theta(\tau_t))$ we recover the on-policy distillation updates used in techniques such as kickstarting~\citep{schmitt2018kickstarting} and Mix\&Match~\citep{czarnecki2018mix}. In this setting there is no corresponding $\alpha_\tau$ that simply rescales policy logits, and thus as a consequence of Theorem \ref{theorem:not_gvf} we see that naive distillation with student-generated trajectories does not form a gradient vector field.  We also note that exactly the same proof shows that the entropy penalty~\citep{mnih2016asynchronous} $\ell(\tau|\theta) = -\sum_{t=1}^{|\tau|}  \text{H}(\pi(\tau_t))$, commonly used in actor critic algorithms, also results in updates that do not correspond to a valid gradient vector field.

Having seen that these commonly used updates do not correspond to a valid gradient vector field, a natural question to ask is whether this is necessarily problematic. For example -- the updates used in Q-learning are not gradient steps either, but Q-learning still provides a convergent iterative scheme. We address the question of what can be said about the dynamical system emerging from this sort of distillation,
and for a simple tabular setup we can show that indeed this is not an issue: 
\begin{proprep}
Using an update rule of the form $\mathbb{E}_{\pi_\theta}[ \sum_{t=1}^{|\tau|} \nabla_\theta \ell(\pi(\tau_t), \pi_\theta(\tau_t)) ]$ for a strongly stochastic\footnote{Meaning that each for each action $a$, parameters $\theta$ and state $s$, $\pi_\theta(s)[a] > 0$.} student policy, with episodic finite state-space MDPs and tabular policies, provides convergence to the teacher policy over all reachable states for the loss function $\ell$, provided the optimiser used can minimise $\ell(a,b)$ wrt. $b$, for any $a$ in the domain of $\ell$, and $\ell(a,b)$ reaches minimum at $\ell(a,a)$.
\end{proprep}
\begin{appendixproof}

Because of strong stochasticity of $\pi_\theta$, the distribution of states visited under this policy covers entire state space $\mathcal{S} = (s_1, \dots, s_N)$ reachable from the initial state. We use notation $\ell^i_\theta := \ell(\pi(s_i), \pi_\theta(s_i))$.
Consequently the update 
$$
g(\theta) := \mathbb{E}_{\pi_\theta}[ \sum_{t=1}^{|\tau|} \nabla_\theta \ell(\pi(\tau_t), \pi_\theta(\tau_t)) ]
$$
can be rewritten as
$$
g(\theta)= \left [ p_\theta(s_1) \nabla_{\theta_1} \ell^1_\theta \;\;\;\;
\dots \;\;\;\;
p_\theta(s_N) \nabla_{\theta_N} \ell^N_\theta
\right ]^T,
$$
where $p_\theta(s)$ is the probability of agent being in state $s$ when following policy $\pi_\theta$ and we use the independence of parametrisation of the policy in each state (which comes from the tabular assumption -- $\theta_i$ is the parametrisation of policy in state $s_i$). 

Let us denote by $g^*(\theta)$ gradient of a an expected loss under teacher policy
\begin{equation*}
\begin{aligned}
g^*(\theta) :&= \nabla_\theta [ \mathbb{E}_{\pi} \sum_{t=1}^{|\tau|} \ell(\pi(\tau_t), \pi_\theta(\tau_t)) ]\\
&= \mathbb{E}_{\pi} \nabla_\theta [ \sum_{t=1}^{|\tau|} \ell(\pi(\tau_t), \pi_\theta(\tau_t)) ]\\
&= 
\left [ p(s_1) \nabla_{\theta_1} \ell^1_\theta \;\;\;\;
\dots \;\;\;\;
p(s_N) \nabla_{\theta_N} \ell^N_\theta
\right ]^T.
\end{aligned}
\end{equation*}
where again $p(s)$ is the probability of sampling state $s$ under $\pi$.

It is easy to notice that these two update directions have a non-negative cosine:
$$
\langle g(\theta), g^*(\theta) \rangle = \sum_{i=1}^N 
 p(s_i)p_\theta(s_i) \| \nabla_{\theta_i} \ell^i_\theta \|^2 \geq 0.
$$
Furthermore, because for all $s$, $p(s)\geq 0, p_\theta(s)>0$, the cosine is zero if and only if for each state $s_i$ either
$
\| \nabla_{\theta_i} \ell^i_\theta \|^2 = 0 
$ (teacher and student policies match)
or $p(s_i)=0$ (state is not reachable by $\pi$).
This means that for every state, reachable by $\pi$, the corresponding update rule coming from $g(\theta)$ is guaranteed to be stricly descending as long as it is not in the minimum.

Due to assumptions about $\ell(a, \cdot)$ having a unique minimum and optimiser being able to find it, we obtain that $\pi_\theta(s_i)$ will converge to $\pi(s_i)$ for each $s_i \in \mathcal{S}$ where $p(s_i)>0$.

Consequently we have shown, that the update direction is a strict descent direction wrt. expected loss under the teacher policy and thus student policy converges to the teacher one over all reachable states.

Using Monte Carlo estimates for the $g(\theta)$ estimation can be analysed analogously to how Stochastic Gradient Descent generalises Gradient Descent.
\end{appendixproof}
\begin{figure*}[htb]
\centering
\begin{tabular}{cc}
\includegraphics[height=6cm]{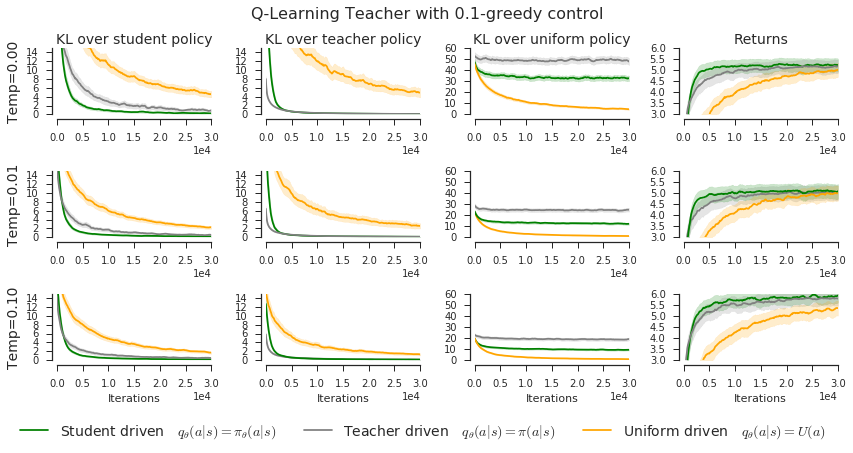}
& 
\includegraphics[height=6cm]{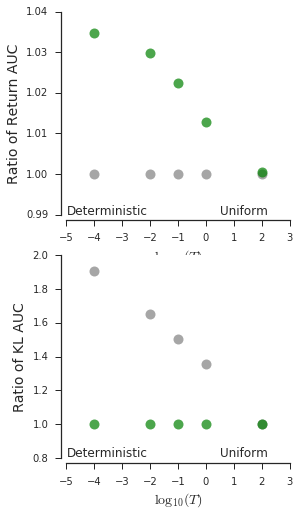}
\end{tabular}
\caption{\textbf{Left:} 
Learning curves averaged over 1k grid worlds when distilling Q-Learning teacher with various control policies (colors) and temperatures (rows). We report KL over various sampling distributions $z_\theta$ (columns) and returns when following teacher policy. Shaded region represents 0.95 confidence intervals, estimated as 1.96 $\cdot$ standard error of the mean.
\textbf{Right:} Relation between student-driven distillation speed-up (measured as ratio of areas under reward/KL curves) and teacher determinism.
}
\label{fig:fig:qu}
\label{fig:entropy}
\end{figure*}
However, despite this positive result, we can also show that even in the episodic MDP case one can break convergence if we introduce rewards. 
The counterexample, visualised in  Fig.~\ref{fig:counter} and described in detail in the Appendix C, relies on a teacher that can discriminate between some states that the student cannot. It leads to an oscillation -- the student policy will never converge even after infinitely many steps. 
\begin{toappendix}
\paragraph{Oscillation example }
Consider a game with seven states, $\{ s_0, s_L, s_R, s_{LL}, s_{LR}, s_{RL}, s_{RR} \}$. We start at $s_0$ and in the first step we decide whether to go to $s_L$ or $s_R$. If we chose to go to $s_L$, in step 2 we chose whether to go to $s_{LL}$ or to $s_{LR}$. Similarly, if we are in $s_R$ after round 1, in step 2 we have a choice whether to go to $s_{RL}$ or $s_{RR}$. The only rewards are $r(L,s_{L}) = -1$, $r(R,s_{L}) = -2$, and $r(R,s_{R}) = -3$. In the game we use a policy $\pi_\theta$ depending on two parameters $\theta_x$ and $\theta_y$ as follows. In the first step we go to $s_R$ with probability $\frac{e^{\theta_x}}{1+e^{\theta_x}}$ and to $s_L$ with probability $\frac{1}{1+e^{\theta_x}}$. In step 2 we have two branchings again, if we are in $s_L$ with probalility $\frac{e^{\theta_y}}{1+e^{\theta_y}}$ we go to $s_{LL}$, and with probability $\frac{1}{1+e^{\theta_y}}$  we go to $s_{LR}$. Similarly, if we are in $s_R$ we go with probalility $\frac{e^{\theta_y}}{1+e^{\theta_y}}$ to $s_{RL}$, and with probability $\frac{1}{1+e^{\theta_y}}$  we go to $s_{RR}$. 
We choose a penalty function $\ell = \ell({\theta_y}) = 4 \frac{e^{\theta_y}}{1+e^{\theta_y}}-4$, living in the state $s_R$, when we are in $s_L$ in step $2$, $\ell$ is zero. Equivalently one can think of it being a distillation cost with an information potential loss, $\ell(\pi(s) \| \pi_\theta(s)) = 4\sum_a \pi(a|s) \pi_\theta(a|s) - 4$ where the teacher $\pi(R|s_L) = 1$.
We have an update rule
\begin{equation*}
\begin{aligned}
&\left\{
\begin{array}{l}
\dot{x} = \frac{\partial}{\partial {\theta_x}}\mathbb{E}_{\pi_\theta}[\sum_{t=1}^{|\tau|} r_t]\\
\dot{y} = \frac{\partial}{\partial {\theta_y}}\mathbb{E}_{\pi_\theta}[\sum_{t=1}^{|\tau|} r_t] - \frac{e^{\theta_x}}{1+e^{\theta_x}} \ell'({\theta_y})\\
\end{array}
\right.\\
&\left\{
\begin{array}{l}
\dot{x} = \frac{e^{\theta_x}(e^{\theta_y}-1)}{(1+e^{\theta_x})^2(1+e^{\theta_y})}\\
\dot{y} = \frac{e^{\theta_y}(1+3e^{\theta_x})}{(1+e^{\theta_x})(1+e^{\theta_y})^2}  -4 \frac{e^{\theta_x} e^{\theta_y}}{(1+e^{\theta_x})(1+e^{\theta_y})^2} 
\end{array}
\right.\\
&\left\{
\begin{array}{l}
\dot{x} = \frac{e^{\theta_x}(e^{\theta_y}-1)}{(1+e^{\theta_x})^2(1+e^{\theta_y})}\\
\dot{y} =  \frac{e^{\theta_y}(1-e^{\theta_x})}{(1+e^{\theta_x})(1+e^{\theta_y})^2}.
\end{array}
\right.
\end{aligned}
\end{equation*}
This system of equations has a first integral $H({\theta_x},{\theta_y}) = e^{\theta_x} + e^{-{\theta_x}} + e^{\theta_y} + e^{-{\theta_y}}$ (with integrating factor $\frac{e^{\theta_x} e^{\theta_y}}{(1+e^{\theta_x})^2(1+e^{\theta_y})^2}$). Note, that $H({\theta_x},{\theta_y}) = 4 + {\theta_x}^2 + {\theta_y}^2 + \mathcal{O}({\theta_x}^3,{\theta_y}^3)$, therefore the fixed point $\theta = (0,0)$ is a center. Therefore, with each policy update the values $\theta$ stay on the same closed curve and they keep changing in a cyclic manner, never converging.
\end{toappendix}

There are multiple possible ways to construct
on-student-policy distillation learning methods similar to the ones used in practice, but which do provide update rules that are gradient vector fields\footnote{It is worth noting that the typical trick of importance sampling is not viable here. First, it is unclear what sampling distribution to correct with respect to -- one can choose any distribution that is independent of $\theta$, thus it could be teacher policy, but also say -- a uniform one. Second, mathematically this simply leads to degeneration to optimisation of the corresponding loss, such as teacher distill rather than on-policy method.}. 
One such way is to start from the objective suggested by the update rule component, namely:
$
\mathbb{E}_{\pi_\theta} \left [ \sum_{t=1}^{|\tau|} \ell(\pi(\tau_t) \| \pi_\theta(\tau_t)) \right ].
$
Then we compute its gradient using the log-derivative trick, analogously to how the KL-regularised RL objective is derived~\citep{schulman2017equivalence} or how Stochastic Computation Graphs are obtained~\cite{schulman2015gradient}; doing so gives the update direction:
\begin{equation*}
\begin{aligned}
\mathbb{E}_{\pi_\theta} \left [\nabla_\theta \ell(\tau|\theta) \right ] + 
\mathbb{E}_{\pi_\theta} \left [\nabla_\theta  \log \pi_\theta(\tau) \ell(\tau|\theta) \right ] .
\end{aligned}
\end{equation*}
As we can see, the gradient vector field is composed of two expectation terms. 
The first term corresponds to the 1-step on-policy distillation setup discussed so far. The second term corresponds to the standard RL objective if $-\ell$ plays the role of the reward function.
This simple derivation allows us to prove the following:
\begin{thmrep}
In order to recover the gradient vector field property for 1-step on-policy distillation updates with any loss $\ell(\pi(\tau_t) \| \pi_\theta(\tau_t))$, one can add an extra reward term $\hat r_t = -\ell(\pi(\tau_{t+1}) \| \pi_\theta(\tau_{t+1}))$. Analogously if the loss is of the form $\mathbb{E}_{a\sim \pi_\theta} \hat \ell(\pi(\tau_t))$ then the correction is of form $-\hat \ell(\pi(\tau_{t+1})).$
\end{thmrep}
\begin{appendixproof}
Consider the following loss
$
\mathcal{L}(\theta) = \mathbb{E}_{\pi_\theta}[ \ell(\tau, \theta) ]
$
and its gradient:
\begin{equation*}
\begin{aligned}
\nabla_\theta\mathcal{L}(\theta) &= \nabla_\theta\int_\tau  \pi_{\theta}(\tau|\theta) \left [ \ell(\tau, \theta) \right ] d\tau \\
&= \int_\tau  \nabla_\theta( \pi_{\theta}(\tau|\theta) \left [ \ell(\tau, \theta) \right ] ) d\tau\\
&=\int_\tau  
[\nabla_\theta\pi_{\theta}(\tau|\theta)] \ell(\tau, \theta)  
+    
\pi_{\theta}(\tau|\theta) [ \nabla_\theta\ell(\tau, \theta) ]  
d\tau 
\end{aligned}
\end{equation*}
using the log-derivative trick $\nabla_\theta f(x) = f(x) \nabla_\theta\log f(x)$ and the above equation we get
\begin{equation*}
\begin{aligned}
\nabla_\theta\mathcal{L}(\theta) &=&& 
\int_\tau  
[\pi_{\theta}(\tau|\theta)  \nabla_\theta\log \pi_{\theta}(\tau|\theta) ] \ell(\tau, \theta)  
+    \\
&&& \;\; \; \; \; \pi_{\theta}(\tau|\theta) [ \nabla_\theta\ell(\tau, \theta) ]  
d\tau \\
&=&& \int_\tau  
[\pi_{\theta}(\tau|\theta)  \nabla_\theta\log \pi_{\theta}(\tau|\theta) ] \ell(\tau, \theta)  
d\tau 
+ \\   
&&&\int_\tau  
\pi_{\theta}(\tau|\theta) [ \nabla_\theta\ell(\tau, \theta) ]  
d\tau \\
&=&& \mathbb{E}_{  \pi_{\theta}(\tau|\theta) } \nabla_\theta\log \pi_{\theta}(\tau|\theta)  \ell(\tau, \theta)  
+   \\ 
&&&\mathbb{E}_{
\pi_{\theta}(\tau|\theta)} \nabla_\theta\ell(\tau, \theta) 
\end{aligned}
\end{equation*}
Consequently, we obtain that the valid gradient of the loss considered is composed of two expectations, one being the equivalent of a RL target, but with $\ell$ being a negation of the reward, and one which is exactly the auxiliary cost of interest. Consequently if we add the reward at time $t$ equal to minus loss at time $t+1$ we will recover proper gradient vector field.

For the case of a loss of the form $\mathbb{E}_{a\sim \pi_\theta} \hat \ell(\pi(\tau_t))$ this proof is analogous -- simply the correction is not on a state-action pair level, rather a pure state level.
\end{appendixproof}

Since this is a gradient vector field, it can be safely composed with reward based updates without losing any convergence properties. As one can see on the right of Fig.~\ref{fig:counter} applying this correction to On-policy distill+R (and thus creating the \emph{N-distill+R}), leads to convergence and minimisation of the loss, as expected.

Given the potential convergence issues with the naive updates from 
Equation \ref{eqn:update_rule}, particularly when also considering reward from the environment as highlighted by our counterexample, it begs the question: 
Why does following the student's policy when performing distillation typically lead to better empirical results?
Our main hypothesis is that, \emph{if convergent\footnote{In practise researchers often force convergence by learning rate annealing, early stopping etc. so the issues highlighted here may often be masked.}, it provides more robust policies wrt. trajectories sampled from the student}.

This follows the general machine learning principle of training in the same regime as that which we expect to encounter during test time. In particular,  after distillation, the goal is usually to either evaluate a student agent when it is generating its own actions, or to allow the student agent to continue training on its own. Therefore what matters is an expectation wrt. $\pi_\theta$, and not wrt. $\pi$. 
Consequently performing distillation ``on-policy'' with respect to student trajectories leads to less of a distribution-shift between training and testing phases. 

Another motivating argument is that if the teacher is almost deterministic,
then it visits a relatively small fraction of the state space, even though during training it might have built a policy to deal with other situations too. 
When using $\pi$ during distillation, the student will not have the opportunity to replicate the teacher's behaviour in these states, since it will not visit them. 
In general, after distillation,  $\pi\neq\pi_\theta$
and they diverge quickly in complex environments or over long trajectories.
Again, on-student-policy distillation avoids these issues, especially initially -- many states will be visited that are not normally encountered under the teacher policy. 
Consequently one can expect better replication of the teacher, when measured in the entire state space.

The main observations of this section are: (i) on-policy 1-step distillation updates do not form gradient fields, and when mixed with environment rewards can lead to non-convergent behaviour; (ii) distillation using student-generated trajectories replicates the teacher policy in more states that are relevant under the student's behaviour distribution.
The following empirical section highlights the effects that different choices for the control-policy have in practice.
\begin{figure*}[htb]
\centering
\includegraphics[width=\textwidth]{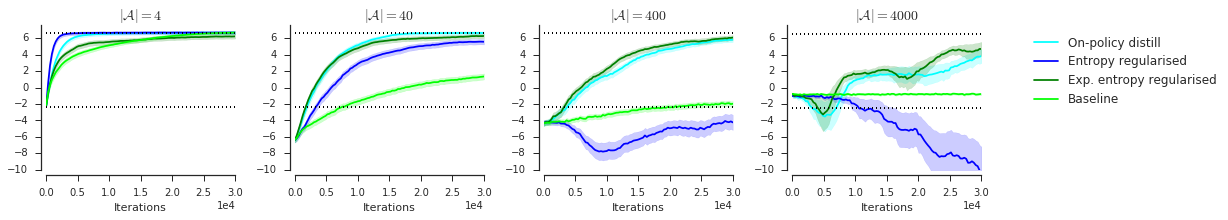}
\caption{Learning curves averaged over 1k random MDPs with $|\mathcal{A}|$ actions, out of which 4 are movement actions and the remaining ones do not affect the movement of the agent, but simply make exploration hard. Plots show the failure mode of the intrinsic reward only based distillation, and how their expected version fixes it.}
\label{fig:actionspace}
\end{figure*}

\subsection{Empirical evaluation of the control policy}
\label{sec:control_policy_emp}

We consider teacher driven~\citep{rusu2015policy}, student driven~\citep{parisotto2015actor} and fixed (uniform) control policies.
We define a distribution over grid world tasks, where we randomly place walls, terminating and rewarding states in 20$\times$20 2D grid worlds
(see Appendix A.2 for a detailed description of the MDP generating procedure), and agents are capable of moving in 4 directions. There is a fixed probability of terminating each episode, such that we end up with bounded (undiscounted) returns. We sample 1k MDPs like this, and distill for 30k optimisation steps using various control policies $q_\theta$.
Since we have proven that in the tabular case we can use distillation based on per-step cross-entropy, H$^\times$, this is the loss we are minimising, using a gradient based update to the underlying logits.

We use teachers trained with Q-Learning and $\epsilon$-greedy policies ($\epsilon$ set to 0.1, full details provided in the Appendix A.4) and observe how different types of control policy affect the distillation loss. We measure
$
\mathbb{E}_{z_\theta} \left [ \sum_{t=1}^{|\tau|} \mathrm{H}^\times(\pi(\tau_t) \| \pi_\theta(\tau_t) ) \right ]
$
for various choices of $z_\theta$, which can be different from $q_\theta$ used for distillation. 

As predicted by our theoretical analysis, student-driven distillation brings benefits wrt. the loss computed over student trajectories. 
In other words, if one cares about how closely the student behaviour matches the teacher behaviour when the student agent is allowed to generate experience on its own, student-driven distillation optimises this quantity well, with the gap disappearing as the teacher is more and more uniform (entropic), see Fig.~\ref{fig:entropy}. Similarly, matching of the teacher policy outside of typically visited states (measured by $z_\theta$ being uniform) is also much better when following a student-driven policy. The best result, in terms of whole state space, is achieved when using uniform distribution for $q_\theta$, but this is a control setting which does not scale to larger state or action spaces. And when we compare this control setting to the student-based setting we see that it converges extremely slowly even in these scenarios. 

Similarly, in terms of the returns obtained by the agent, student driven distillation sees the fastest learning progress, and needs on average 3$\times$ less steps than teacher driven distillation (and around 10$\times$ less than uniform driven) to recover full teacher performance. This is of crucial importance, since distillation in RL is typically a first step in a larger training procedure in which the rewards obtained by the student policy are to be maximised with potentially further training. 
In order to be useful for these applications, one typically seeks to obtain highly rewarding policies as rapidly as possible. The only criterion under which teacher driven distillation works more effectively is, somewhat obviously, the expected KL under trajectories generated from teacher distribution. However this is an artificial scenario, which is rarely encountered in practice.

To summarise, student-driven distillation provides significant improvements in terms of empirical results over teacher-driven distillation. While one could heuristically drive the switch between the two~\cite{ross2011reduction}, the pure student-driven method seems to be strong enough to use solely, as long as a proper loss is being used, which we discuss in the next section.
Therefore, in the remainder of this paper we focus on student driven distillation $q_\theta = \pi_\theta$.

\subsection{Empirical evaluation of various updates}
\label{sec:control_loss_emp}
An important choice is the selection of which method we use to update the student policy given the trajectory and actions suggested by the teacher policy. There are two popular approaches here: one is to try to maximise the probability of the trajectory generated by the student under the teacher policy~\citep{teh2017distral,schulman2017equivalence}, and the other is to frame the learning as a per-timestep supervised learning problem, defining the loss at each timestep to be the cross entropy between the teacher's and student's distributions over actions~\citep{arora2018multi,schmitt2018kickstarting,czarnecki2018mix}

There are two aspects worth discussing. Firstly, in the entropy regularised setup, where we use $\text{H}^\times(\pi_\theta\|\pi)$ the student would be considered a prior, while the teacher a posterior, however in the distillation setup with $\text{H}^\times(\pi\|\pi_\theta)$ the teacher is the prior and the student the posterior.
Secondly, the cross entropy regularised approach (which tries to minimise the cross entropy between whole trajectories distributions $\mathbb{E}_{\pi_\theta}[\sum_t - \log \pi(a_t|\tau_t)]$) 
can be absorbed in the reward channel, without any $\ell$ being used. Using only the reward signal can suffer from very high variance in the gradient estimator, for example when the action space is large.
As one can see from Fig.~\ref{fig:actionspace}, as we increase the size of the action space from four actions up to 4k actions (by adding many actions that do not move the agent) the speed of entropy regularised distillation drastically collapses, while traditional distillation still works well. 

\begin{figure*}[htb]
    \centering
    \includegraphics[width=\textwidth]{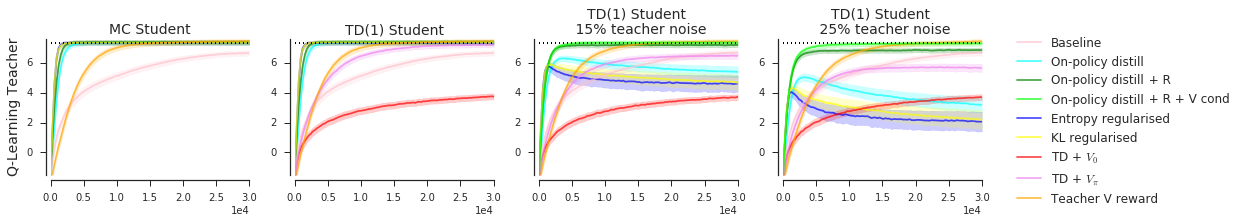}
    \caption{Learning curves obtained from averaging 1k training runs on randomly generated MDP grid worlds. The gradual decrease in reward when distilling from a sub-optimal Q-Learning teacher with 
    distillation methods that enforce full \emph{policy cloning} comes from the fact that the teacher is purely deterministic -- while being closer to it initially helps, once the student replicates all the wrong state decisions perfectly its reward start to decrease.}
    \label{fig:big_comp}
\end{figure*}
One can reduce the variance by splitting the entropy term into a 1-step update expressed by $\ell$, and incorporate the remaining updates through $\hat r$. This technique, denoted as \emph{expected entropy regularised}, indeed recovers performance of traditional distillation. While this is not a new objective as such, but rather a different estimator, it is, to our best knowledge a novel method, which strictly dominates popular alternatives.

The direction chosen for the cross entropy has a very simple intuitive explanation. If one uses $\text{H}^\times(\pi\|\pi_\theta)$ then one tries to replicate $\pi$, however if one uses $\text{H}^\times(\pi_\theta\|\pi)$ then one tries to find a deterministic policy, which puts all probability mass on the most probable action of $\pi$ (see the Appendix C for a proof).
If the cost is changed to be the KL divergence instead, this issue is eliminated for finite action spaces, however for continuous control it is a matter of mean vs median seeking techniques~\cite{minka2005divergence} (see the Appendix C for more detail). Depending on the MDP, both methods can be beneficial, and of course for almost deterministic teachers -- they actually are equivalent.

To summarise, these experiments demonstrate that across the space of possible student-driven distillation approaches the most reliable method, both mathematically and empirically, is our proposed \emph{expected entropy regularised distillation}. 
It has three key benefits: (i) it creates a valid gradient vector field; (ii) it does not suffer from high-variance of the estimate typically used in similar methods~\citep{teh2017distral}; and (iii) it directly maximises the probability of the student produced trajectories under the teacher policy, as opposed to the \emph{n-distill} method, which looks at maximising the probability of being in states where the student and teacher agree. Consequently, it combines the best elements of various similar techniques in a single method, and avoids their respective drawbacks.

\begin{toappendix}
\paragraph{Cross entropy minima}
Let us fix a distribution $p(a|s)$, and consider a minima of $\text{H}^\times(p\|q)$ and $\text{H}^\times(q\|p)$ wrt. $q$. It is easy to see that the minimum of $\text{H}^\times(p\|q)$ is given by $p$, as by the definition of divergence, the minimum of $\text{KL}^\times(p\|q)$ is given by $p$, and $\text{KL}^\times(p\|q) = \text{H}^\times(p\|q) + \text{H}(p)$, but for a fixed $p$, $\text{H}(p)$ is a constant, thus it does not affect the minima. For $\text{H}^\times(q\|p)$ we will show that the minimum is given by the dirac delta distribution in the most probable action $a^*$ in $p$, denoted as $q^*$. For simplicity, assuming that this is a unique action, meaning that $\forall_{a \neq a^*} p(a|s) < p(a^*|s)$, then for any $q \neq q^*$
\begin{equation*}
\begin{aligned}
\text{H}^\times(q\|p) &= - \sum_a q(a|s) \log p(a|s)\\
&> 
- [ \sum_a q(a|s) ] \max_b \log p(b|s)\\
&= - [ 1 ] \log p(a^* | s) = \text{H}^\times(q^*\|p)
\end{aligned}
\end{equation*}

\begin{figure}[h]
    \centering
    \includegraphics[width=0.15\textwidth]{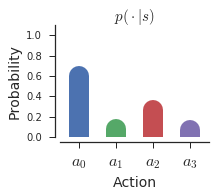}
    \includegraphics[width=0.15\textwidth]{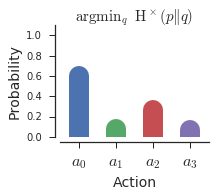}
    \includegraphics[width=0.15\textwidth]{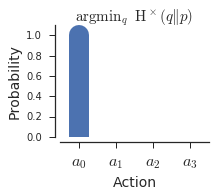}
    \caption{Comparison of various cross-entropies solutions when matching the distribution over finitely many actions.}
    \label{fig:hs}
\end{figure}
\end{toappendix}

\begin{toappendix}
\paragraph{KL and mean/mode seeking}
While both $\mathrm{KL}(q\|p)$ and $\mathrm{KL}(p\|q)$ have the same minimum in the space of all distributions, they differ once one constrains the space we are looking over. To  be more precise we have that
$$
\arg\min_q \mathrm{KL}(q\|p) = \arg\min_q \mathrm{KL}(p\|q) = p
$$
but at the same time there exists $C \subset \mathcal{P}$ where $\mathcal{P}$ is the space of all distributions such that
$$
\arg\min_{q \in C} \mathrm{KL}(q\|p) \neq \arg\min_{q \in C} \mathrm{KL}(p\|q) \neq p
$$
\begin{figure}[h]
    \centering
    \includegraphics[width=0.23\textwidth]{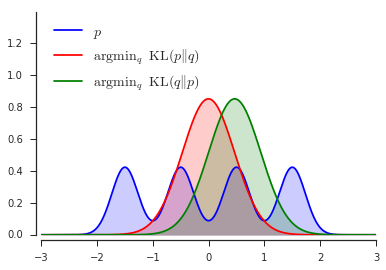}
    \includegraphics[width=0.23\textwidth]{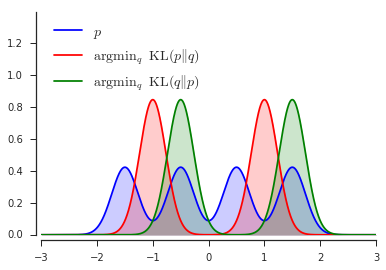}
    \includegraphics[width=0.23\textwidth]{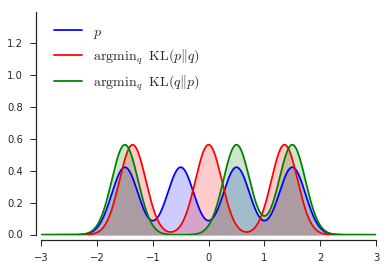}
    \includegraphics[width=0.23\textwidth]{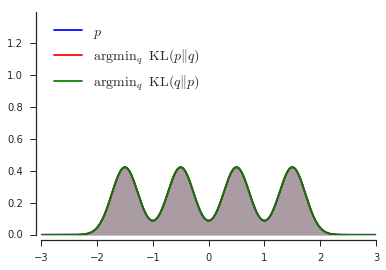}
    \caption{Comparison of various KL variant solutions when matching the distribution over a mixture of 4 Gaussians, using from 1 (upper left) to 4 (lower right) Gaussians. Note how mode seeking KL (green) picks Gaussians to match, while ignoring others, and mean seeking (red) instead puts its Gaussians in between peaks of the original distribution.}
    \label{fig:kls}
\end{figure}
\begin{figure}[h]
    \centering
    \includegraphics[width=0.23\textwidth]{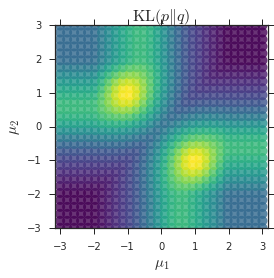}
    \includegraphics[width=0.23\textwidth]{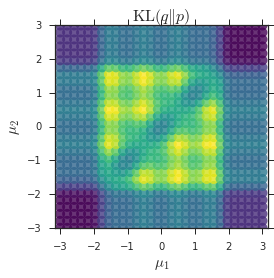}
    \caption{Visualisation of the value of KL (left) and reverse KL (right) parameterised by the location of the mean of two Gaussians, computed with respect to the mixture of 4 Gaussians from Fig.~\ref{fig:kls}. One can see how mean seeking KL prefers to put means in -1 and 1 while the mode seeking attains the minimum for every possible pair, matching means of the original mixture.}
    \label{fig:kls2}
\end{figure}
The simplest example is the mixture of multiple Gaussians, which we try to fit with just a single Gaussian. The typical cost of $\mathrm{KL}(p\|q)$ will match the mean of the distribution (thus the name of \emph{mean seeking}), while $\mathrm{KL}(q\|p)$ will cover one of the Gaussians from the mixture, while ignoring the others (thus \emph{mode seeking}), see Fig.~\ref{fig:kls} and Fig.~\ref{fig:kls2}.

In practice, we are often in this regime, since the teacher and student policies can have different capacities, architectures and priors, thus making perfect replication impossible. Therefore, the choice of \emph{direction} of KL will affect if the agent prefers to just match one, very probable mode (action/behaviour), or if we prefer the agent to look for an averaged action/behaviour.

\end{toappendix}

\section{Policy distillation from Actor-Critic}
\label{sec:actor_critic}

In practice, while pure policy based methods remain useful~\citep{ross2011reduction} in RL, the Actor-Critic framework~\citep{mnih2016asynchronous, schulman2015trust,espeholt2018impala} has risen to prominence in recent years. 
Consequently we now shift our attention towards distillation strategies that make use of value critics, denoted by $V_\pi(s) = \mathbb{E}_\pi[ \sum_t r_t | \tau_1 = s]$. The availability of this additional knowledge, in the form of $V_\pi(s)$, allows us to better leverage imperfect teachers, as we can begin to estimate how much to trust them.

For example, let's assume that we have ground-truth access to $V_{\pi_\theta}$ and $V_{\pi}$, and consider a loss of the form:
$
\ell(\theta, s) = \text{H}^\times( \pi(s) \| \pi_\theta(s) ) [V_\pi(s) - V_{\pi_\theta}(s) ]_{>0} ,
$
where $[x]_{>0} = 1$ iff $x>0$. This can be seen as an action-independent version of the Generalised Policy Improvement technique~\cite{barreto2017successor}.
As a result we can easily prove that after converging, our agent will be at least as strong as its teacher, independent of the initial returns of $\pi_\theta$. 
\begin{proprep}
For $\mathcal{S}_1$ being a distribution over initial states, 
if we have $\forall s \in \mathcal{S} \; \ell(\theta^*,s) \leq \text{H}( \pi(s) )$ then
$\mathbb{E}_{s \sim \mathcal{S}_1} [ V_{\pi_{\theta^*}}(s) ] \geq
\mathbb{E}_{s \sim \mathcal{S}_1} [V_\pi(s)] $.
\label{prop:gate}
\end{proprep}
\begin{appendixproof}
Lets assume that the inequality does not hold, meaning that the following teacher's policy gives higher return. This means, that there exists a state $s^*$, where $V_\pi(s^*) > V_{\pi_{\theta^*}}(s^*)$  but the policies differ, meaning that $\pi(s^*) \neq \pi_{\theta^*}(s^*)$. However, if $V_\pi(s^*) > V_{\pi_{\theta^*}}(s^*)$ then $\ell(\theta, s^*) = \text{H}^\times( \pi(s^*) \| \pi_\theta(s^*) )$, and due to the assumption $\ell(\theta, s^*) \leq \text{H}( \pi(s^*) )$ for every state, leads to $\pi(s^*) = \pi_{\theta^*}(s^*)$ (as cross entropy is equal to entropy of the first argument only when the argument are the same), which is a contradiction.
\end{appendixproof}

With techniques like this we can use teachers which do make mistakes. One such experiment, where we randomly flip the most probable action of the teacher policy in a given percentage of states it visits, is illustrated in Fig.~\ref{fig:big_comp}. As before we generate 1k random MDPs, and empirically estimate $V_\pi$ by sampling 100 trajectories after a given fraction of state-action pairs have been modified.
As one can see, methods that fully replicate the teacher (in this case a Q-Learning based one, with 25\% noise) end up saturating in sub-optimal solutions -- around 2-4 points (Fig.~\ref{fig:big_comp} blue and cyan curves). If we add true rewards to the learning system, we improve, but still saturate around 6.5 points (Fig.~\ref{fig:big_comp} dark green curve). With the value function based gating described in Proposition \ref{prop:gate} we recover the full performance of the teacher with approximately 7 points (Fig.~\ref{fig:big_comp} lime curve). As expected, the usefulness of this approach depends on the quality of the teacher and the accuracy of the value function estimators.

\subsection{Using the teacher's critic to bootstrap}
Yet another way of using the teacher's knowledge is to use its value functions, $V_\pi$, instead of the student's, $V_{\pi_\theta}$, for bootstrapping -- so that the usual actor-critic TD(1) update direction:
$ \nabla \log \pi_\theta(a_t|\tau_t) [ r(a_t, \tau_t) + V_{\pi_\theta}(\tau_{t+1}) ] $
instead becomes
$
\nabla \log \pi_\theta(a_t|\tau_t) [ r(a_t, \tau_t) + V_{\pi}(\tau_{t+1}) ] ,
$
which we will refer to as \emph{$\mathrm{TD}+ V_\pi$}.
Note, that it is no longer optimising for the true return
since the teacher's value function is not with respect to the student's policy.
Consequently, it is not obvious if the resulting update is a gradient vector field,
nor whether the result of the updates has meaningful fixed points.
We can show that for simple cases it is a valid update rule:
\begin{proprep}
Assume that we are given the true value $V_\pi$ of the teacher policy $\pi$, for a finite state size MDP, then optimising using 
$
\mathbb{E}_{\pi_\theta}[ \sum_t \nabla \log \pi_\theta(a_t|\tau_t) [ r(a_t, \tau_t) + V_{\pi}(\tau_{t+1}) ] ].
$
converges to a policy with $\mathbb{E}_{s \sim \mathcal{S}_1} V_{\pi_\theta}(s) \geq \mathbb{E}_{s \sim \mathcal{S}_1} V_\pi(s)$ for $\mathcal{S}_1$ being the distribution of initial states.
\end{proprep}
\begin{appendixproof}
First, notice that for all initial states, the update rule provided basically solves the bandit problem, where the value of each action is a sum of an actual reward and the value of the teacher (implying following the teacher policy afterwards). In the worst case scenario it will simply find a distribution matching the teacher's, as it is going to optimise for the reward in the first step, and then fall back to the teacher's policy. 
Consequently, after enough updates, the policy $\pi_\theta$ will learn to take actions which do not have smaller values than those of the teacher if one was to follow the teacher policy afterwards. 
Now, using inductive reasoning, if $\pi_\theta$ is already defining a distribution over states visited up to $n$ steps from the initial state which are guaranteed to produce values larger than the teacher, and if we were to follow teacher policy afterwards, then the update will also correct states in distance $n+1$. Given that we assumed that it is a finite state size MDP and updates to different states are independent, then the whole process has to eventually converge.
\end{appendixproof}
As the previous experiments show (Fig.~\ref{fig:big_comp}), bootstrapping from the teacher leads to a better performance than just replicating it (which empirically confirms the theoretical claim of an improvement), but at the same time it is not solving the original problem, and so the method will still saturate early on, if the teacher is not strong enough (e.g.: see the very poor results of $\mathrm{TD} + V_0$ where the teacher just predicts 0 value everywhere).

\subsection{Using the critic as an intrinsic reward}
\label{sec:int_reward}

Another possible way to use the teacher's critic is to define an intrinsic- (or shaping-) reward based on its value
$
\hat r^V_t := V_\pi(\tau_{t+1}) - V_\pi(\tau_{t}) + r_t,
$
which will provide reward proportional to the increase in value over the last action. Note, that now we have to add true reward, as value can decrease when the rewarding state is encountered and we do not want to penalise our agent for actually obtaining the reward.

It is easy to conclude, that such an update rule is going to guarantee convergence to the policy which has the maximal value, as:
\begin{equation*}
    \begin{aligned}
&\mathbb{E}_{\pi_\theta} [ \sum_{t=1}^{|\tau|} \hat  r^V_t] =
\mathbb{E}_{\pi_\theta} [ \sum_{t=1}^{|\tau|} V_\pi(\tau_{t+1}) - V_\pi(\tau_{t}) +  r_t] 
\\&= 
\mathbb{E}_{\pi_\theta} [  - V_\pi(\tau_1)  + \sum_{t=1}^{|\tau|} r_t ]
= 
\mathbb{E}_{\pi_\theta} [  \sum_{t=1}^{|\tau|}  r_t ] + \text{const.}
    \end{aligned}
\end{equation*}
while at the same time it is not forcing complete policy cloning even if $\pi$ is the optimal policy (which is the property of any potential based intrinsic reward~\citep{ng1999policy,asmuth2008potential,devlin2012dynamic}).

Although the presence of such a value-function based shaping reward does not affect the optimal solutions achieved, the learning dynamics are affected. 
If the teacher is strong (close to optimal) then it is going to 
help convergence speed; if it is weak (obtains low returns), it can slow down training.
We can formalise it with the following proposition
\begin{proprep}
Let us assume the teacher is an optimal policy for the given MDP, then for each action $a_t$ that would lead to a deviation from the optimal path, it will get an immediate penalty, meaning that $r^V_t < r_t$, while following any of the optimal paths leads to $r^V_t = r_t$.
\label{prop:optimal_teacher}
\end{proprep}
\begin{appendixproof}
It is easy to notice, that if an agent executes an action $a_t$ which is on the optimal path, we have $V_\pi(\tau_{t+1}) = V_\pi(\tau_{t})$, and thus $r^V_t =  V_\pi(\tau_{t+1}) - V_\pi(\tau_{t}) + r_t = 0 + r_t = r_t$.
If, on the other hand, it is not on the optimal path, then there exists $\epsilon >0$ such that $ V_\pi(\tau_{t+1}) = V_\pi(\tau_{t}) - \epsilon$ so  $r^V_t =  V_\pi(\tau_{t+1}) - V_\pi(\tau_{t}) + r_t = -\epsilon + r_t < r_t$
\end{appendixproof}

In a symmetric counterpoint to Proposition \ref{prop:optimal_teacher}, if our teacher is the worst possible policy (minimising returns rather than maximising) we get the opposite proposition -- where one receives immediate penalties for doing anything but the worst possible actions. While this still does not affect the solution of learning, it will slow down training significantly.

As a concrete example of these concepts, we can use a simple grid world where one can only go left or right, while starting in the middle of a corridor of length $2T+1$, and rewards +1 and -1 are placed at the right and left ends of the corridor respectively. We define an optimal teacher $\pi_\text{opt}(R|s_i) = 1.0$ and an adversarial teacher $\pi_\text{bad}(L|s_i) = 1.0$ (but we set $V_{\pi_\text{bad}}(a|s) := V_{\pi_\text{good}}(a|s)$, thus creating adversarial value function).
\begin{figure}[]
    \centering
    \includegraphics[width=0.5\textwidth]{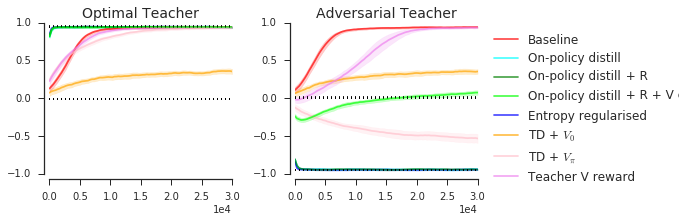}
    \caption{Results of distilling an optimal teacher and an adversarial (minimising reward) teacher from the chain-structured MDP described in Section~\ref{sec:int_reward} with $T=5$. Analogous results for other $T$ are provided in Appendix Fig.~\ref{fig:mdp_v_example_ext}. }
    \label{fig:mdp_v_example}
\end{figure}
When distilling from an optimal policy, distillation techniques based on policy cloning 
recover optimal scores very quickly (Fig.~\ref{fig:mdp_v_example}).
Knowledge transfer using shaping rewards from the teacher's value function
leads to much slower progress, but it still outperforms our baseline training without any distillation.
However, once we switch to the adversarial teacher, all the policy cloning approaches fail, while the \emph{Teacher V reward} still learns.
It is also worth noting that bootstrapping from the teacher's value function fails in this task, as it is only guaranteed to improve upon the teacher (which it does), not to solve the original task.

\section{Conclusions}

In this paper we sought to highlight some of the strengths, weaknesses, and potential mathematical inconsistencies in different variants of distillation used for policy knowledge transfer in reinforcement learning. In particular, we suggest a unifying view of many different techniques which allows them to be compared and understood side-by-side. We provide both theoretical analyses, as well as large scale empirical studies on synthetic MDPs, which allows us to provide a prescriptive suggestion of best-practices for distillation in different settings. The synthesis of these findings is summarised in the flowchart in Figure \ref{fig:main}.

A key contribution of our work is to demonstrate that many widely used methods do not correspond to valid gradient vector fields, and thus may be susceptible to non-convergent learning dynamics. 
However, armed with our insights, we are able to suggest modifications to these approaches which address some of these dynamical issues.
In particular we found that expected entropy regularised distillation seem to be the most 
reliable formulation of distillation, both theoretically and based on our empirical results. Also, if available, the critic of the teacher policy can be used to deal with imperfect teachers.

While the synthetic MDPs used in this work were crucial for exploring effectively different hypothesis, as future work, an open question is how these approaches behave empirically on large-scale, real-world problems, using function approximators like deep neural networks. Mathematically understanding the role of the function approximator is also left as future work.

\bibliographystyle{plain}
\bibliography{arxiv}

\end{document}